\documentclass{vgtc}
\ifpdf
\pdfoutput=1\relax 
\usepackage{graphicx}
\DeclareGraphicsExtensions{.pdf,.png,.jpg,.jpeg} 
\else
\usepackage{graphicx}
\DeclareGraphicsExtensions{.eps} 
\fi%

\graphicspath{{figures/}{pictures/}{images/}{./}} 

\usepackage{microtype} 
\PassOptionsToPackage{warn}{textcomp}
\usepackage{textcomp}
\usepackage{mathptmx}
\usepackage{times} 
\usepackage{cite}
\usepackage{tabu}
\usepackage{booktabs} 
\usepackage{amsmath}
\usepackage{multirow}
\usepackage{diagbox}
\usepackage{array}
\usepackage{float}
\usepackage{color,colortbl}
\definecolor{Gray3}{gray}{0.9}
\definecolor{Gray2}{gray}{0.7}
\definecolor{Gray1}{gray}{0.5}
\usepackage{tabularx,colortbl}
\usepackage{algpseudocode}
\usepackage{algorithm}

\onlineid{0}

\vgtccategory{Research}
\vgtcpapertype{please specify}

\title{DroneARchery: Human-Drone Interaction through Augmented Reality with Haptic Feedback and Multi-UAV Collision Avoidance Driven by Deep Reinforcement Learning}

\author{Ekaterina Dorzhieva\thanks{e-mail: ekaterina.dorzhieva@skoltech.ru} %
\and Ahmed Baza \thanks{e-mail: ahmed.baza@skoltech.ru} %
\and Ayush Gupta\thanks{e-mail: ayush.gupta@skoltech.ru} %
 \and Aleksey Fedoseev\thanks{e-mail: aleksey.fedoseev@skoltech.ru} %
 \and Miguel Altamirano Cabrera\thanks{e-mail: miguel.altamirano@skoltech.ru} %
 \and Ekaterina Karmanova\thanks{e-mail: ekaterina.karmanova@skoltech.ru} %
 \and Dzmitry Tsetserukou\thanks{e-mail: d.tsetserukou@skoltech.ru} %
 }
\affiliation{\scriptsize Skolkovo Institute of Science and Technology, Russia}

\shortauthortitle{Dorzhieva \MakeLowercase{\textit{et al.}}: DroneARchery: Human-Drone Interaction through Augmented Reality with Haptic Feedback and Multi-UAV Collision Avoidance Driven by Deep Reinforcement Learning}
 \abstract{We propose a novel concept of augmented reality (AR) human-drone interaction driven by RL-based swarm behavior to achieve intuitive and immersive control of a swarm formation of unmanned aerial vehicles. The DroneARchery system developed by us allows the user to quickly deploy a swarm of drones, generating flight paths simulating archery. The haptic interface LinkGlide delivers a tactile stimulus of the bowstring tension to the forearm to increase the precision of aiming. The swarm of released drones dynamically avoids collisions between each other, the drone following the user, and external obstacles with behavior control based on deep reinforcement learning.
 
 The developed concept was tested in the scenario with a human, where the user shoots from a virtual bow with a real drone to hit the target. The human operator observes the ballistic trajectory of the drone in an AR and achieves a realistic and highly recognizable experience of the bowstring tension through the haptic display. 
 
 The experimental results revealed that the system improves trajectory prediction accuracy by 63.3\% through applying AR technology and conveying haptic feedback of pulling force. DroneARchery users highlighted the naturalness (4.3 out of 5 point Likert scale) and increased confidence (4.7 out of 5) when controlling the drone. We have designed the tactile patterns to present four sliding distances (tension) and three applied force levels (stiffness) of the haptic display. Users demonstrated the ability to distinguish tactile patterns produced by the haptic display representing varying bowstring tension(average recognition rate is of 72.8\%) and stiffness (average recognition rate is of 94.2\%). 
 
 The novelty of the research is the development of an AR-based approach for drone control that does not require special skills and training from the operator. In the future, the proposed interaction can be applied in various fields, for example, for fast swarm deployment in search and rescue missions, crop monitoring, inspection and maintenance. 
}
\CCScatlist{
  \CCScatTwelve{Human-centered computing}{Human computer interaction (HCI)}{Interaction devices}{Haptic devices};
  \CCScatTwelve{Human-centered computing}{Human computer interaction (HCI)}{Interaction paradigms}{Mixed / augmented reality}
}

\teaser{
\centering
\includegraphics[width=\linewidth]{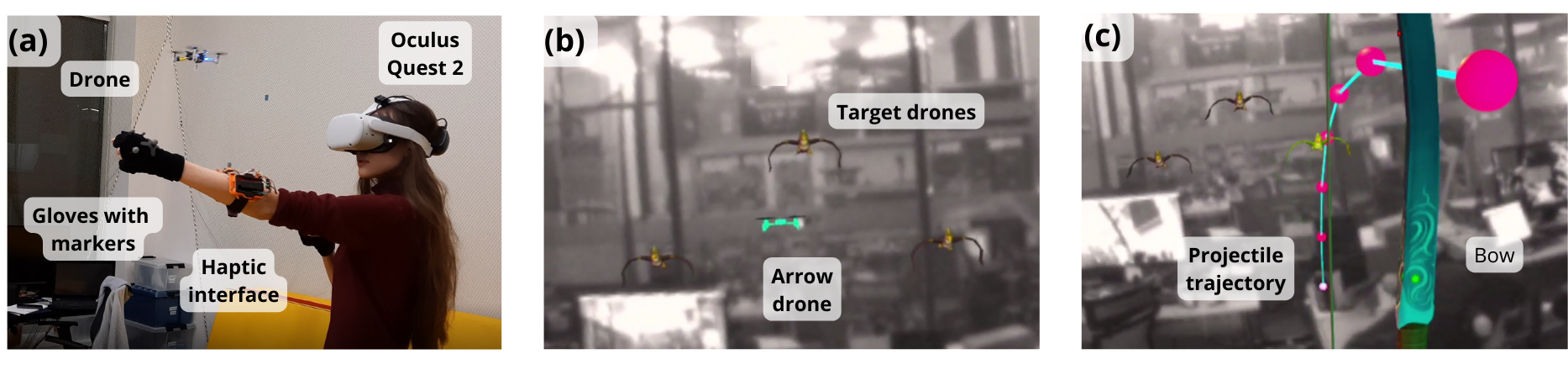}
\caption{(a) Interaction between human wearing haptic interface and drones in-game scenario. (b) Real “arrow" drone and virtual “target" drones in augmented reality (AR) environment. (c) The ballistic trajectory of the “arrow" drone is rendered via visual and haptic interfaces to the user.}
\label{fig:teaser}
}

\begin{document}


\firstsection{Introduction}

\maketitle

The role of human-robot interaction (HRI), in virtual and augmented reality applications, is becoming increasingly important. In industry, medicine, and entertainment, robots are mainly utilized for either supporting humans or performing tasks in a shared space with an operator. For example, BitDrones by Gomes et al. \cite{gomes2016bitdrones}, introduced a novel human-drone interaction (HDI) paradigm where nano-quadrotors are used as self-levitating tangible building blocks, forming an interactive 3D display with a touchscreen array. 


Currently, users are faced with the challenges of controlling a swarm of drones. Joystick-based control may be feasible only for a single agent, while more complex systems require highly interactive solutions \cite{2019}. A novel system SwarmCloak introduced by Tsykunov et al. \cite{SwarmCloak} suggests the landing of a swarm of four nano-quadrotors on the light-sensitive pads with vibrotactile feedback, attached to the user's arms.
Robot control systems require user training, which can be carried out during game interaction. Novel games are bringing more attention to human-drone interactions, driving the development in this field. With the involvement of more users in the HDI game process, completely new ideas and solutions appear. SwarmPlay suggests a novel HDI scenario when a swarm of drones plays against the human \cite{swarmplay}.  

This paper explores tactile and visual feedback interfaces that make the user experience with drones more immersive and intuitive. Several researchers explored haptic interaction with robots in virtual reality, where collaborative robots \cite{Fedoseev_2020}, mobile robots and drones \cite{Abdullah_2018} generate tactile stimuli. Moreover, the VR applications with robots are explored in the scope of post-stroke \cite{Clark_2019} and upper limb \cite{Luo_2020} rehabilitation. Additionally, human-centered virtual and augmented reality applications were also extensively investigated for the purpose of entertainment \cite{Juhasz_2008}. 

Deep learning methods are currently actively used in robotics, e.g, for gesture recognition in the teleoperation tasks \cite{Montebaur_2020}. To use the optimal number of agents during land surveying and swarm communication, it is necessary to continuously adjust the configuration of the swarm depending on the data received from the sensors. 

In this paper, we propose a new approach for rapid deployment of a drone swarm, including intuitive, accurate, and safe control through the integration of a haptic display, the use of AR, and Deep Reinforcement Learning (DRL) based collision avoidance. The experiments were conducted to verify the performance of DroneARchery system. The first experiment (Sec. \ref{sec:exp1}) explores the ability of a person to distinguish between signals given by a haptic display. The second experiment (Sec. \ref{sec:exp2}) examines the accuracy of single drone control according to the presented concept and its usability for users regardless of their experience with drones. The third experiment (Sec. \ref{sec:exp3}) is aimed at evaluating the DRL method for avoiding obstacles when deploying a swarm of drones using DroneARchery technology.

\section{Related Works}
\subsection{Haptic Displays in Human-Robot Interaction Scenarios}

The need for an intuitive and efficient human-robot interface has led to several applications based on haptic interfaces, allowing to inform the operator of the robot state and its surroundings through tactile cues \cite{Kavas_2018, Duan_2019}. 

Many state-of-the-art approaches to control a single drone or a swarm rely on a wearable tactile interface, such as an intuitive system by Robeiro et al. \cite{Ribeiro_2018} that is composed of two small and lightweight wearable displays to control UAV for racing competitions. The wearable displays preserving the high mobility of the human operator were proposed by Byun et al. \cite{Byun_2019}, where epidermal tactile sensor arrays were applied to achieve the direct teleoperation of the swarm by human hand.

Previously developed systems have achieved high precision in the haptic rendering of contact forces, however, they do not cover many scenarios involving UAV control. In this research, we propose an HDI concept, in which the haptic display is rendering the string tension of the bow for drone ballistic trajectory generation. In contrast to the previously suggested tactile interfaces, this research explores the capabilities of a multimodal display to intuitively and accurately convey a sense of elastic tension. The effectiveness of pressure and displacement feedback on the user's perception of tension in an elastic body is also investigated in this work (Sec. \ref{sec:exp1}).

In recent research, there is an increasing interest in delivering haptic feedback to the user's forearm to decrease the restriction of their motion. Moriyama et al. \cite{8357173} proposed a multi-contact interface delivering a haptic interaction experience to the forearm by two inverted five-bar linkage mechanisms inspired by the LinkTouch design introduced by Tsetserukou et al. \cite{Tsetserukou_2014}. The combination of two contact points allowed the researchers to render shear and normal forces with a high recognition rate. The application of this device in archery games, however, is estimated to be less effective as the direction of linkage motion does not correlate with a sling or bow tension. Another approach is explored by Shim et al. \cite{10.1145/3491101.3519908} with the QuadStretch wearable display, where skin stretching of the forearm is achieved by the relative displacement of two bracelets. One of the applications suggested for QuadStretch includes a VR archery game, where the bow tension is transmitted as the magnitude of stretching force. In DroneARchery, we relied on a combination of linear displacement of a single contact point and normal forces applied by the LinkTouch display placed collinearly to the bow string displacement. Our hypothesis was that the linear displacement may provide highly distinguishable feedback about the bow tension, which was later supported by the experimental results.

\subsection{Virtual and Augmented Reality Interfaces}

VR and AR interfaces have been utilized extensively for immersive human-robot interaction and game scenarios. Ibrahimov et al. \cite{ibrahimov2019dronepick} developed a VR-based teleoperation system DronePick that performs delivery of objects by a human-controlled quadrocopter. The virtual interface of DronePick supports the selection of target objects and real-time control over the drone position. VR interfaces are capable of solving complex problems of operator presence in many work processes. For example, Kalinov et al. \cite{kalinov2021warevr} proposed an interface to supervise an autonomous robot remotely from a secluded workstation in a warehouse. 

With the high improvement in AR hardware, several projects have investigated the ability of AR to improve the control over a swarm of drones \cite{Liu_2020, Macchini_2020}. A multi-channel robotic system for human-swarm interaction in augmented reality is presented by Chen et al. \cite{Chen_2020}. 

While drone in-flight control through AR is being extensively explored, the deployment and docking of large-scale swarms remain to be addressed. 

\subsection{Swarm Collision Avoidance}

Obstacle avoidance is a fundamental part of the operation of unmanned vehicles. Several probabilistic algorithms are widely implemented in multi-agent systems, e.g., Rapidly-exploring Random Tree \cite{RRT} algorithm for calculating an obstacle-free path for the swarm of several UAVs. 
DRL improves problem-solving with non-linear function approximation \cite{9564258}. The DRL approach in the shooting was explored by Nikonova and Gemrot \cite{DBLP:journals/corr/abs-1910-01806}, in which the trained agent made the most successful shot that scores the most points, while we oppositely train the agents to avoid obstacles by dodging the shot.


This paper suggests the DRL algorithm for collision avoidance in the drone swarm during the “arrow" drone deployment, when the high number of “target" drones leads to higher risks of collision. Therefore, we implemented the DRL algorithm to mitigate this problem and evaluated the system performance.

The contribution to the DRL scenarios is that the developed system switches between the real and virtual swarms of drones depending on the user's task. Thus, users in AR can release real drones one by one, which will be launched on site. If the user is working with the swarm remotely or testing the system performance, the rest of the swarm will operate virtually, handled by the same algorithm. Therefore, the same collision avoidance approach is required to work both with real drones and in simulation. Moreover, the simulation serves as an important part of system verification before real launches. In the user study, a virtual swarm of “target" drones interacted with a real “arrow" drone to evaluate the scenario of the remote swarm deployment for operation in hazardous areas.

\section{System Architecture}
\begin{figure}[!h]
 \centering
 \includegraphics[width=1.0\linewidth]
{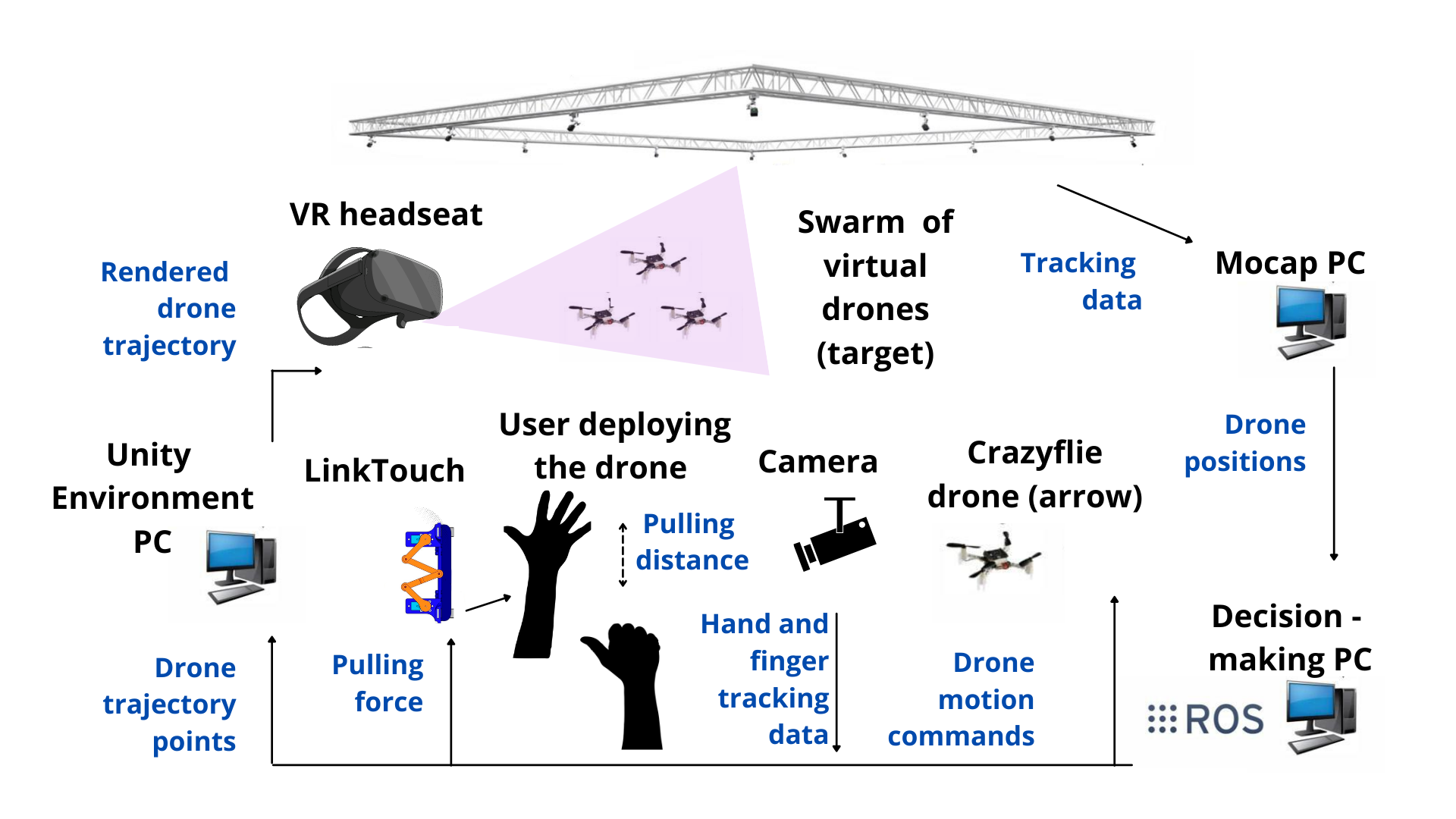}
 \caption{Overview of the DroneARchery system. The user interacts with real (arrow) and virtual (target) drones in an augmented environment and generates real drone trajectory via a gesture interface processed by the CV algorithm. To improve the user aiming we propose to deliver the haptic stimuli representing bow tension to the forearm. The approach of haptic rendering is similar to the sliding bar of computer-based scenarios.}
\label{fig:Overwiev}
\end{figure}

The system architecture in Fig. \ref{fig:Overwiev} presents how the game works through the integration of all system elements. 

The VICON motion capture system tracks the user's palms with attached spherical markers. The acquired hand position is then cued to the ROS framework with the Vicon-bridge package. The user initiates the interaction by holding the virtual bow with one hand and pulling the arrow with the other. The relative hand positions are used to calculate the drone trajectory, based on the virtual bow model described in Sec. \ref{sec:traj}. Further, the trajectory points are provided to the AR environment and the haptic display described in Sec. \ref{sec:AR} and Sec. \ref{sec:display}. The drone follows the user's bow-holding hand and hovers at 15 $cm$ above it, providing a more realistic sense of aiming. We use Crazyflie 2.0 quadcopter (weight is 27 grams, flight time is of 7 minutes). Users move their hands, aiming at the different targets. At the same time, the VR headset provides visual trajectory, and the haptic display renders tactile patterns corresponding to the bow tension calculated from the palms' positions. The hand tracking approach described in Sec. \ref{sec:camera} detects the user unclenching the fingers holding the arrow and sends a signal to the drone to follow the generated parabolic trajectory. Virtual drones in the Unity environment react to the approach of a real drone and avoid the collision thanks to DRL Sec. \ref{sec:RL}.

\subsection{AR Game Environment}
\label{sec:AR}

Several options of environment rendering are being extensively implemented in the AR solutions: video see-through (VST) displays, optical see-through (OST) displays, waveguides, and laser beam scanning. In the DroneARchery scenario, the former rendering method was applied. Thus, the user wears a VR headset Oculus Quest 2, and the cameras capture the world around the user and transmit the processed video to the eyes.  

AR environment is developed through the Passthrough API Experimental technology in Quest 2 Software Development Kit (SDK), which the OpenXR backend can access in Unity. In the AR environment, the user can observe the trajectory of the archery before the shot (Fig. \ref{fig:teaser}(b)). The AR gives the user an immersive experience and helps to hit the target more accurately.

\subsection{Haptic Display}
\label{sec:display}
While users can aim naturally by directing their hands at the target, they have no natural means to estimate the pulling force as it is calculated entirely virtually, and the goal becomes unattainable. Delivering the experience of the tension in the bow to the user poses a challenging task, as the bowstring changes its position and force magnitude at the contact point dynamically. Thus, for our research, we applied a haptic display based on the LinkTouch design, that is attached to the user's forearm. 

\begin{figure}[!h]
 \centering
 \includegraphics[width=0.7\linewidth]{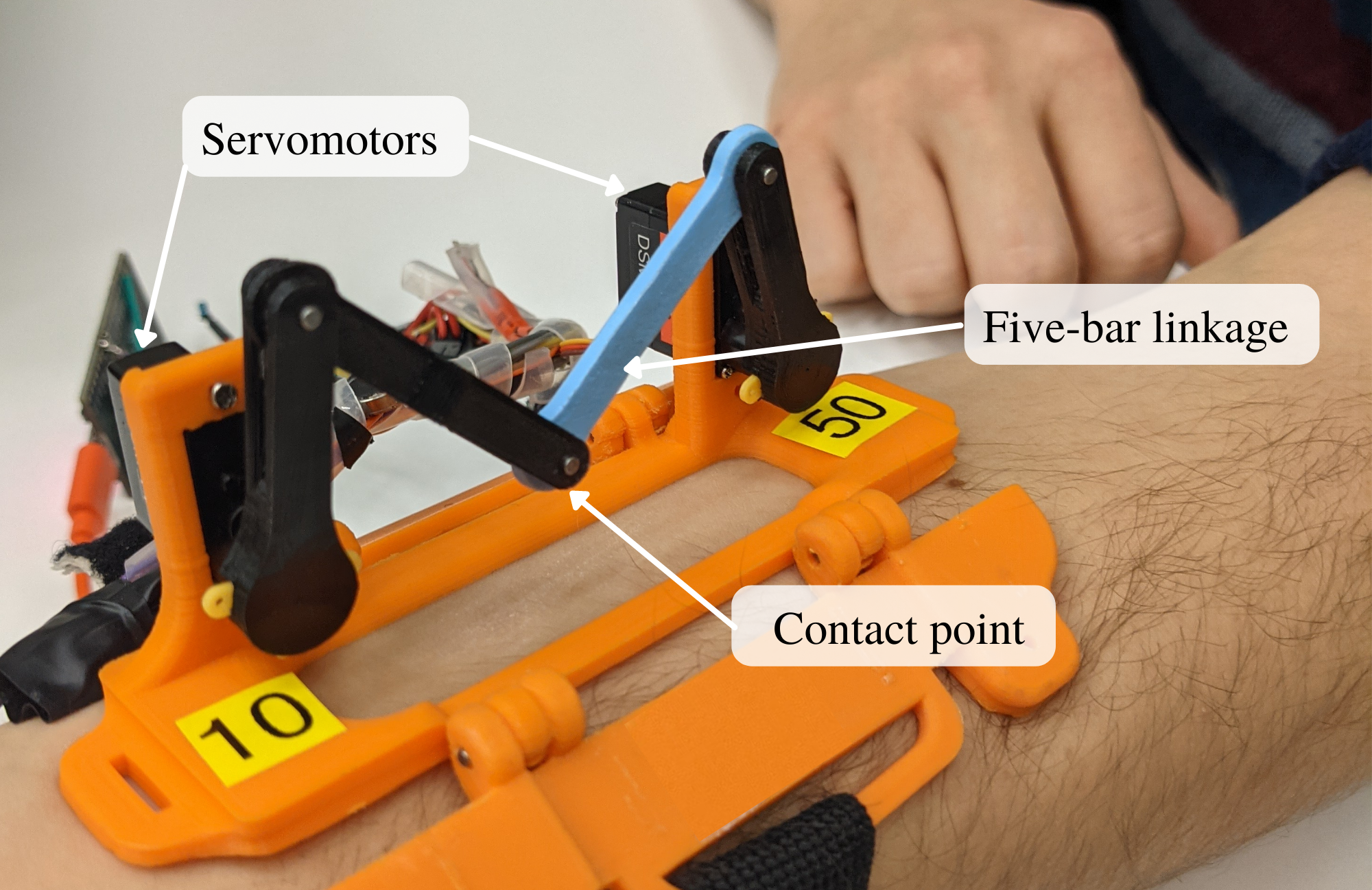} 
 \caption{LinkTouch display with M-shaped linkage for delivering multi-modal (force and position) tactile stimuli.}
 \label{fig:display}
\end{figure}

To provide haptic feedback of virtual elastic string motion, the LinkTouch display was modified and adapted to the user's forearm area (Fig. \ref{fig:display}). LinkTouch is a wearable haptic display with inverted five-bar linkages. This display includes two servomotors that determine the position of the endpoint that touches the user's arm. Thus, the display is capable of both executing a linear motion along the forearm and increasing the pressing force by its motion in the normal direction to the arm. The maximum normal force at the contact point is equal to 2 N. LinkTouch is controlled by an ESP32 microcontroller, which defines the angles of servomotors and handles the Bluetooth communication with the computer. In addition, the low weight of the display (135 grams including the battery) allows the user not to experience physical fatigue, which is confirmed in Sec. \ref{sec:exp2_res}.

By locating the haptic display at the forearm, we considered the sensitivity of the forearm. The directional sensitivity perception at the forearm is six times lower than at the fingertips \cite{doi:10.3109/08990229109144725}, which is acceptable because of the larger dimension of the forearm. Moreover, it was considered that the user’s hands have to be free to emulate the movements of the archer and to be tracked by the CV algorithm.

The known distance between the user's hands is transformed into the coordinates of the contact point generated by the haptic display. We assume that the maximum tension of the arrow corresponds to one meter between palms (this value is averaged and depends on the individual characteristics of each bow model). The operating range of LinkTouch is 75 mm. Based on the data obtained, we convert the distance between the user's palms into the coordinates of the contact point of the mechanism. Thus, clenched palms and palms spread one meter apart correspond to the two extreme positions of the haptic display. In addition to moving along the forearm, the contact point can also go up and down perpendicular, creating different pressure on the skin. In our case, the pressure represents the bowstring stiffness. After the shot, the display moves to the starting position, in which there is no contact with the skin. 

\subsection{Ballistic Trajectory Calculation}
 \label{sec:traj}

The drone trajectory was calculated with the position of the user's hands provided by VICON Vantage V5 motion capture system. The pulling force of the virtual bow was calculated assuming a uniform Hooke's law constant $K$. Thus, the potential energy stored in the bow is given in Eq. (\ref{eq:1}):
\begin{equation} 
 \label{eq:1}
U = K \cdot x^2 , 
\end{equation}
where $U$ is the stored potential energy, and $x$ is the stretching distance of the bowstring. The projectile motion was applied along the vector connecting the user's hands. The projectile initial velocity $v$ from the bow was then calculated from the kinetic energy via Eq. (\ref{eq:2}) at the start of the trajectory derived from the potential energy at the point of release.

\begin{equation}
 \label{eq:2}
K_E = \tfrac{1}{2} \cdot m \cdot v^2 , 
\end{equation}

\noindent where $K_E$ is the kinetic energy of the drone, $m$ is the drone mass.
The resulting velocity vector is defined by:

\begin{equation} 
 \label{eq:4}
\vec{V} =v \cdot \frac{ \vec{P}_{bow}-\vec{P}_{arrow} }{\left \| \vec{P}_{bow}- \vec{P}_{arrow}\right \|}, 
\end{equation}

\noindent where $P_{arrow}$ and $ P_{bow}$ are the positions of the user's hands holding the virtual arrow and bow, $\vec{V}$ is the initial velocity vector. The trajectory of the drone is calculated as projectile motion in 3D
space as described in Eq. (\ref{eq:5}-\ref{eq:7}):

\begin{equation} 
 \label{eq:5}
x = v_0\cdot t \cdot cos(\theta) \cdot sin (\gamma)
\end{equation}

\begin{equation} 
 \label{eq:6}
y = v_0\cdot t \cdot cos(\theta) \cdot cos(\gamma)
\end{equation}

\begin{equation} 
 \label{eq:7}
z = v_0 \cdot t \cdot sin(\theta) - \frac{1}{2} \cdot g \cdot t^2,
\end {equation} 
\noindent where $\theta$ and $\gamma$ are the initial launch angles in the vertical and horizontal planes, $t$ is the time moment, $g =9.8$ $\ m/s^2$ is the acceleration of gravity.

\subsection{Hand Tracking Approach}
 \label{sec:camera}
\begin{figure}[!h]
 \centering
 \includegraphics[width=0.9\linewidth]{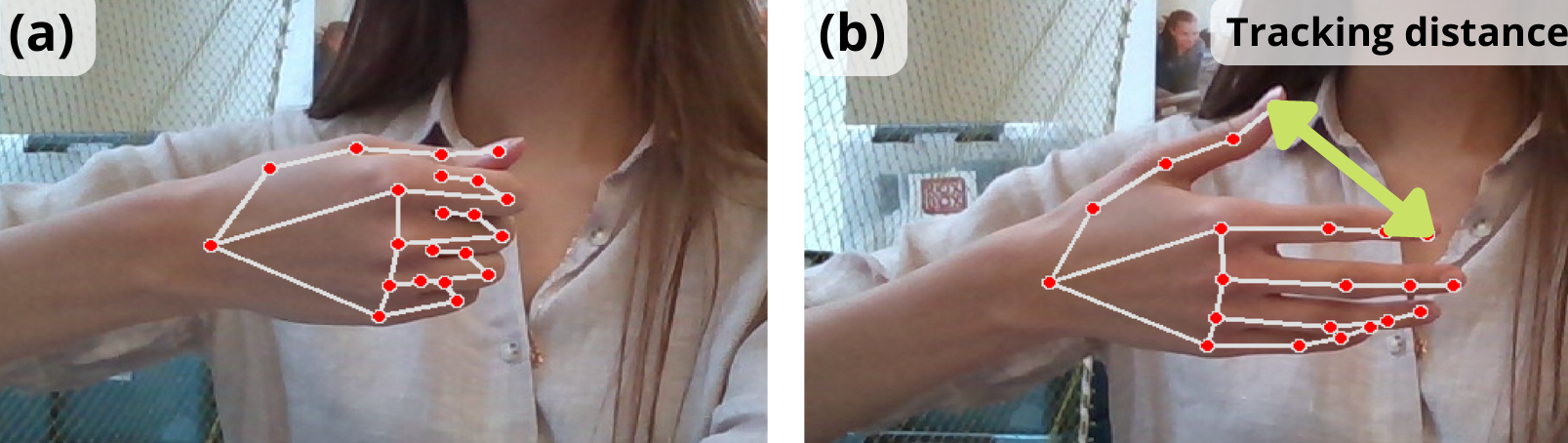} 
 \caption{Finger tracking by MediaPipe framework. (a) Clenched fingers trigger the calculation of a bowstring tension. (b) Unclenched fingers trigger the bowstring release.}
  \label{fig:camera}
\end{figure}

We implemented the computer vision (CV) module to track the user's hand that holds a virtual bowstring. During the game, the drone is switching between two behavior strategies. When users pull the bowstring, the drone follows their left hand with the virtual bow to indicate the launching position. After releasing the bowstring, the drone follows the calculated trajectory. The moment of switching between these two modes is activated by pinching and opening the distance between the fingers holding the bowstring (see Fig. \ref{fig:camera}).

Mediapipe Holistic framework \cite{48292} was applied for the finger tracking module to Logitech HD Pro C920 webcam data for string release estimation. At this moment, the next drone flies up to the user's hand and the procedure is repeated.

\subsection{Deep Reinforcement Learning-based Approach for Collision Avoidance}
\label{sec:RL}


In the game scenario, the “arrow" drone moves at high velocity towards the swarm hovering in formation. The task of the swarm is to avoid a collision with the arrow and each other. The state is a matrix that contains information about the positions of all agents and the position of the arrow in 3D space at each moment. Actions are the velocities that agents develop to reach the next state. Actions are sampled from 3D continuous action space, limited by the drone speed of 0.5 m/s. The reward system was designed for the agents as the sum of the reward for avoiding internal collision between agents given in Eq. (\ref{eqREW:2}) and the reward for formation control given in Eq. (\ref{eqREW:4}).



\begin{equation}
 \label{eqREW:2}
\centering
 r_c =
\begin{cases}
0 & d_i > r_{emergency} \\
- 1 & \forall d_i \leq r_{emergency}
\end{cases}
, 
\end{equation}
\noindent where $d_i$ is the distance between the rewarded agent and the $i$ object in the environment (other drones, an arrow, and borders), $r_{emergency}$ is the critical distance between the agent and object that indicates a high probability of collision of drones.

\begin{equation}
 \label{eqREW:4}
 \centering
 r_f = \begin{cases}
 0 & d_f < r_{formation} \\
 0.01 & d_f \geq r_{formation}
 \end{cases}
, 
\end{equation}

\noindent where $d_f$ is the distance between the current position of the rewarded agent and its target position, $r_{formation}$ is the radius in which the agent is considered to be in the formation.





\begin{figure}[!h]
 \centering
 \includegraphics[width=1.0\linewidth]{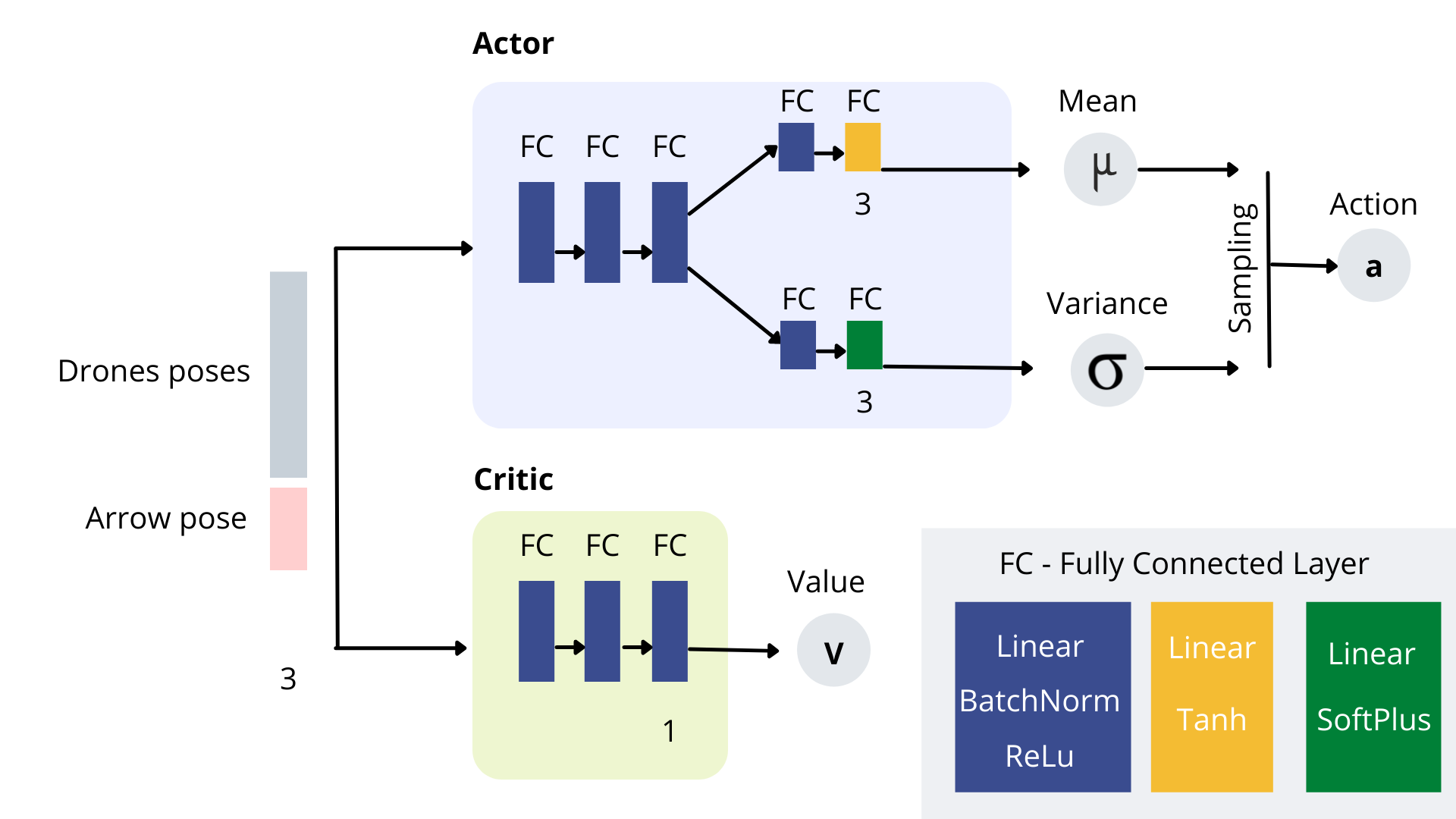}
 \caption{In the Actor-Critic architecture, the state of the environment is input, the Actor-Network calculates the optimal policy, and the Critic-Network calculates the value function.}
\label{fig:network}
\end{figure}

To solve the problem of collision avoidance, the multi-agent Actor-Critic (A2C) approach was applied due to its ability to learn simultaneously the policy and value functions. Each agent has actor and critic networks (Fig. \ref{fig:network}). A2C takes observations as input, then passes them through critics' networks to evaluate how good the current state is, and the actors get a recommendation on what action to take. Actions sample from Gaussian Distribution for each agent. The optimal value was calculated with the Bellman equation for the value function.


The loss of the critic network was calculated as the mean squared error of optimal values and current values. The log probability of an action is given in Eq. (\ref{eqrl:4}).  

\begin{equation}
 \label{eqrl:4}
 \log_{\pi\theta}(a|s) = \sum_{n=1}^{k}(-\frac{(a-\mu_i)^2}{2\sigma_i^2} - \log\sqrt{2\pi\sigma_i^2})(V^*_i(s_t) - V_{\pi}^{U_i}(s_t)),
\end{equation}
\noindent where $k$ is the dimension of action space, $\mu_i$ and $\sigma_i$ are the mean and the variance of policy, respectively, $V^*_i(s_t)$ is the optimal value, calculated with Bellman Equation, and $V_{\pi}^{U_i}(s_t)$ is the current value obtained from critic network with weights $U_i$ of the $i$ agent.


To encourage agents to explore the environment, entropy is subtracted from the actor loss function as in Eq. (\ref{eqrl:6}). Then we update the actor network by minimizing actor loss.
\begin{equation}
 \label{eqrl:6}
ActorLoss=\frac{\sum{\log_{\pi\theta}(a|s)}}{batch size} + \beta (-\frac{\sum{\frac{\log(2\pi\sigma_i^2) + 1}{2}}}{batch size}),
\end{equation}

\noindent where $\beta$ is the hyperparameter that controls the influence of entropy loss, $batchsize$ is the number of states for training.

\section{User Study}
We conducted a two-stage user study to identify the desirable haptic patterns able to provide a pulling experience to users and evaluate the improvement of user performance in a drone-based game scenario with haptic feedback of a virtual bow and with haptic feedback and visualized trajectory. 

\subsection{Tactile Pattern Recognition}
\label{sec:exp1}

\subsubsection*{Participants}
We invited 10 participants, four females and six males, aged 21 to 28 years (mean = 24.2, std = 1.83), to experiment on tactile pattern recognition. According to the laboratory's internal regulations, the participants were informed about both experiments and agreed to the consent form.

\subsubsection*{Procedure} 
\begin{figure}[!h]
 \centering
 \includegraphics[width=0.9\linewidth]{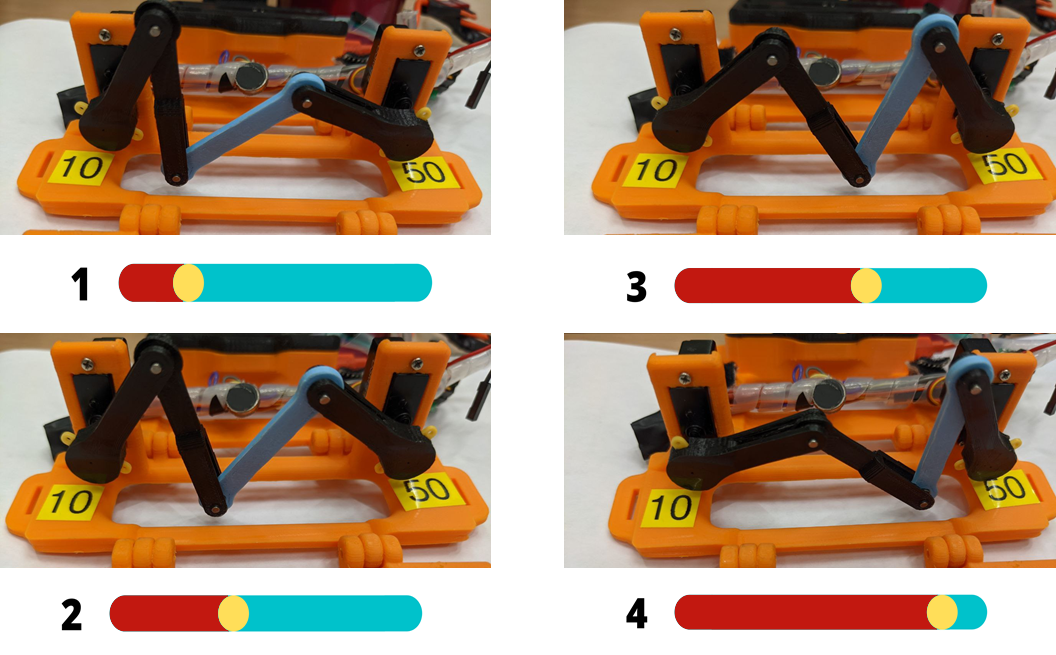}
 \caption {Proposed tactile patterns for distance presentation. The traveled distance by the end effector of the LinkTouch with inverted five-bar linkage is represented by the color bar below.}
 \label{fig:patterns}
\end{figure}

To accurately deploy a swarm of drones with low delays without special training of the operator we propose to use haptic feedback. It delivers to the user the information about the flight path of the drone, same as an archer receives information about the flight path of an arrow from the tension of the bowstring. In this experiment, we explore the haptic display's ability to transmit distinguishable signals to the user. Twelve tactile patterns have been designed to render the distance between the user's palms (Fig. \ref{fig:patterns}) and the pulling force (bowstring stiffness) during the game. Each pattern was designed as a combination of four sliding distances (1-4) and three normal forces applied to the user's forearm (strong as S, medium as M, and light as L).

Before the test, the 12 tactile patterns were demonstrated to the user with comments indicating distance and applied linear forces. The sequence was demonstrated several times according to the user's requirements. Then visual and sound feedback from the haptic display was separated from the user by headphones with white noise and an opaque screen. The haptic display generated tactile patterns in a random order three times (overall 36 patterns were presented to the user) with a 10-second time delay between each pattern. 

\subsection{Experimental Results}

The results of the user study of the tactile patterns applied by LinkTouch were summarized in a confusion matrix shown in Table \ref{tab:conf_matrix}.

\begin{table*}[!ht]
\centering
\caption{Confusion Matrix for Recognition of Pulling Distance Pattern }
\label{tab:conf_matrix}
\scalebox{0.95}{
\begin{tabular}{| c | c | c | c | c | c | c | c | c | c | c | c | c |}
\hline
\multicolumn{1}{|c|}{$\%$} &\multicolumn{12}{c|}{\textit{Predicted}}\\
\hline
\textbf{} Real & 1S & 2S & 3S & 4S & 1M & 2M & 3M & 4M & 1L & 2L & 3L & 4L \\
\hline

1S & \cellcolor{Gray1}\textbf{66.7} & \cellcolor{Gray3}\textbf{13.3} & \cellcolor{Gray3}\textbf{13.3} & 0.0 & \cellcolor{Gray3}\textbf{6.7} & 0.0 & 0.0& 0.0 & 0.0 & 0.0 & 0.0 & 0.0 \\\hline

2S & \cellcolor{Gray3}\textbf{6.7} &\cellcolor{Gray2}\textbf{63.4} & \cellcolor{Gray3}\textbf{23.3} & \cellcolor{Gray3}\textbf{3.3}& 0.0 & \cellcolor{Gray3}\textbf{3.3} & 0.0& 0.0 & 0.0 & 0.0 & 0.0 & 0.0 \\\hline

3S & 0.0 & \cellcolor{Gray3}\textbf{10.0}& \cellcolor{Gray1}\textbf{73.4} & \cellcolor{Gray3}\textbf{13.3}& 0.0 & 0.0 & \cellcolor{Gray3}\textbf{3.3}& 0.0 & 0.0 & 0.0& 0.0 & 0.0\\\hline

4S & 0.0 & \cellcolor{Gray3}\textbf{3.3} & \cellcolor{Gray3}\textbf{6.7} & \cellcolor{Gray1}\textbf{83.3}& 0.0 & 0.0 & 0.0& \cellcolor{Gray3}\textbf{6.7} & 0.0 & 0.0& 0.0 & 0.0\\\hline

1M & 0.0 & 0.0 & 0.0 & 0.0& \cellcolor{Gray2}\textbf{40.0} & \cellcolor{Gray3}\textbf{26.8} & \cellcolor{Gray3}\textbf{3.3}& 0.0 &\cellcolor{Gray3}\textbf{13.3}& \cellcolor{Gray3}\textbf{13.3} & \cellcolor{Gray3}\textbf{3.3} & 0.0 \\\hline

2M & 0.0 & 0.0 & 0.0 & 0.0 & 0.0 & \cellcolor{Gray2}\textbf{56.7} & \cellcolor{Gray2}\textbf{40.0} & \cellcolor{Gray3}\textbf{3.4}& 0.0& 0.0 & 0.0 & 0.0\\\hline

3M & 0.0 & 0.0 & 0.0& 0.0 & 0.0 & \cellcolor{Gray3}\textbf{3.3} & \cellcolor{Gray1}\textbf{80.1} & \cellcolor{Gray3}\textbf{13.3} & 0.0 & 0.0 & \cellcolor{Gray3}\textbf{3.3} & 0.0\\\hline

4M& 0.0& 0.0& 0.0 & 0.0 & 0.0 &0.0& \cellcolor{Gray3}\textbf{20.0} & \cellcolor{Gray1}\textbf{73.3} & 0.0 & 0.0 & 0.0 & \cellcolor{Gray3}\textbf{6.7}\\\hline

1L & 0.0 & 0.0 & 0.0 & 0.0& \cellcolor{Gray3}\textbf{3.3}& 0.0 & 0.0&0.0& \cellcolor{Gray1}\textbf{76.7}& \cellcolor{Gray3}\textbf{16.7} & \cellcolor{Gray3}\textbf{3.3}& 0.0\\\hline

2L & 0.0 & 0.0& 0.0 & 0.0 & 0.0 &0.0 &0.0& 0.0& \cellcolor{Gray3}\textbf{3.3}& \cellcolor{Gray1}\textbf{73.4} & \cellcolor{Gray3}\textbf{23.3} & 0.0\\\hline

3L & 0.0 & 0.0 & 0.0& 0.0& 0.0& \cellcolor{Gray3}\textbf{3.3} & 0.0 & 0.0 & 0.0 & \cellcolor{Gray3}\textbf{10.0} & \cellcolor{Gray2}\textbf{60.0} & \cellcolor{Gray3}\textbf{26.7}\\\hline

4L & 0.0 & 0.0 & 0.0& 0.0& 0.0& 0.0 & \cellcolor{Gray3}\textbf{3.3} & 0.0 & 0.0 & \cellcolor{Gray3}\textbf{3.3} & \cellcolor{Gray3}\textbf{13.3} & \cellcolor{Gray1}\textbf{80.1}\\\hline

\end{tabular}
}
\end{table*}

To evaluate the statistically significant difference between the perception of the patterns, we analyzed the results using single-factor repeated measures ANOVA (the normal force was set to be in linear proportion to the sliding distance in this experiment), with a chosen significance level of $\alpha<0.05$. According to the ANOVA results, there is a statistically significant difference in the recognition rates for the different combinations of applied forces and sliding distances, $F(11,108) = 1.936, p = 0.038$. Two-way ANOVA showed no significant interaction effect between tension and sliding distance value in user recognition: $F = 2.71, p_{int} = 0.15 > 0.05$. The experiment results revealed that the average pattern recognition rate is $68.9$\%. The highest recognition for all forces was achieved with the endpoints of display sliding ( $83.3$\% in position 4 for high force, $80.1$\% in position 4 for low force).
However, the overall recognition of patterns paired with the applied forces is revealed to be higher, as shown in Table \ref{tab:force}. Results of tactile pattern recognition paired by the distance (Table \ref{tab:distance}) show that long-distance was the most discernible to users, and short distances were more often (in 23.4\% and 28.9\%) mistaken for longer ones.

\begin{table}[!ht]
\centering
\caption{Confusion Matrix of Force Recognition}
\label{tab:force}
\scalebox{0.95}{
\begin{tabular}{| c | c | c | c |}
\hline
\multicolumn{1}{|c|}{ $\%$} &\multicolumn{3}{c|}{\textit{Estimated force}}\\
\hline
\textbf{}Applied force & S & M & L \\
\hline
S & \cellcolor{Gray1}\textbf{95.0} & \cellcolor{Gray3}\textbf{5.0} & 0.0 \\\hline
M & 0.0 &\cellcolor{Gray2}\textbf{90.0} & \cellcolor{Gray3}\textbf{10.0} \\\hline
L & 0.0 & \cellcolor{Gray3}\textbf{2.5}& \cellcolor{Gray1}\textbf{97.5} \\\hline

\end{tabular}
}
\end{table}

\begin{table}[!ht]
\centering
\caption{Confusion Matrix of Distance Recognition}
\label{tab:distance}
\scalebox{0.95}{
\begin{tabular}{| c | c | c | c | c |}
\hline
\multicolumn{1}{|c|}{ $\%$} &\multicolumn{4}{c|}{\textit{Estimated distance}}\\
\hline
\textbf{}Distance & 1 & 2 & 3 & 4\\
\hline
1 & \cellcolor{Gray1}\textbf{68.9} & \cellcolor{Gray2}\textbf{23.4} & \cellcolor{Gray3}\textbf{7.7} & 0.0 \\\hline
2 & 3.3 &\cellcolor{Gray1}\textbf{65.6} & \cellcolor{Gray2}\textbf{28.9} & \cellcolor{Gray3}\textbf{2.2} \\\hline
3 & 0.0 & \cellcolor{Gray3}\textbf{8.9}&\cellcolor{Gray1}\textbf{73.4} &\cellcolor{Gray3}\textbf{17.8}\\\hline
4 & 0.0 & \cellcolor{Gray3}\textbf{2.2}&\cellcolor{Gray3}\textbf{14.4} & \cellcolor{Gray1}\textbf{83.4} \\\hline

\end{tabular}
}
\end{table}

The average recognition rate of the forces is $94.2\%$, which supports the hypothesis behind LinkTouch that tactile feedback may be applied to improve the perception of elastic tension. 

\subsection{Experiment on Conveying the Tangible Experience of Virtual Archery Game}
\label{sec:exp2}
 
\subsubsection*{Participants}
This experiment involved 28 users, twelve females and sixteen males, aged 22 to 27 years (mean = 23.93, std = 1.94); several used had the experience of interacting with AR interfaces and drones (13 users). They were formed into four groups with an equal distribution of participants by gender and experience with the drone swarm. Before starting the experiment, the participants were given time to familiarize themselves with the system and take test shots to minimize individual differences in the learning effect.

\subsubsection*{Procedure} 
\begin{figure}[h]
 \centering
\includegraphics[width=0.8\linewidth]{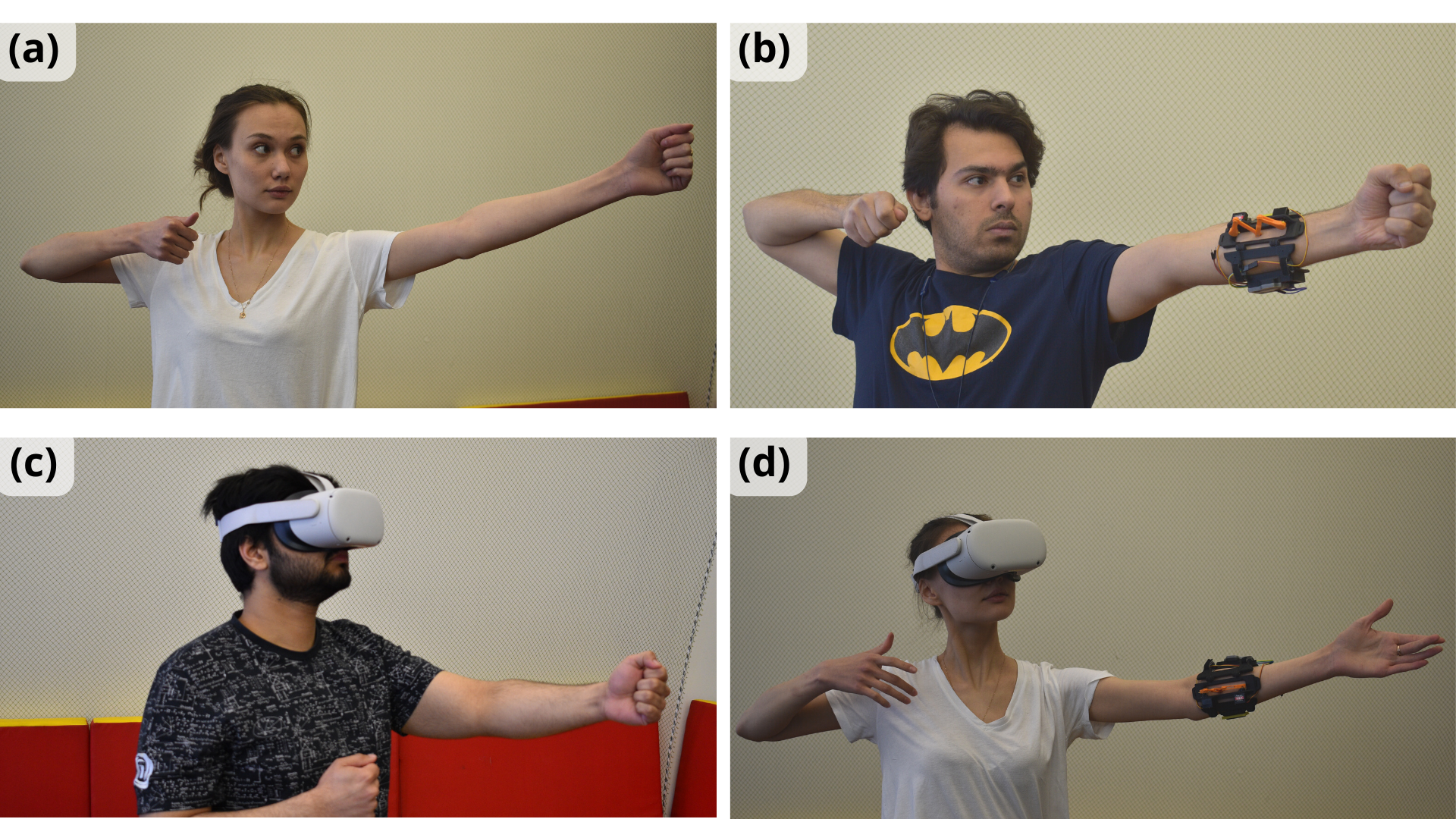}
 \caption{Four groups of participants play DroneARchery. (a) The users are in a real environment without the visualized trajectory and tactile feedback (R group). (b) The users using haptic feedback from LinkTouch display without the visualized trajectory (RH group). (c) The users perform with AR interface without LinkTouch display (AR group). (d) The users perform with both the AR interface and LinkTouch display (ARH group).}
 \label{fig:users}
\end{figure}

The aim of the experiment was to earn the maximum number of points in the archery game. The task could be accomplished by deploying the drone from a virtual bow in 3 gates of different sizes. The scoring policy was as follows: in each of the 3 gates, it was required to fire 3 shots, if the shot was successful, the user earned 1, 3, or 5 points according to the size of the gate, thus, the maximum number of points was equal 27. A shot was counted as successful when the drone was hitting a gate (Fig. \ref{fig:users}).

Before the start of the game, a drone flew up towards the user's left hand and then followed the hand until the moment of the shot with a shift along the z-axis of 15 cm. The users of the first evaluation group were aiming at the gates in a real environment without the visualized drone's trajectory and tactile feedback (R group). The participants of the second group were using haptic feedback from LinkTouch display (RH group) and the third group was performing with Oculus Quest 2 headset (AR group). The last fourth group was allowed to use both visual and tactile displays (ARH group). 
The ARH and AR groups could observe the flight path of the drone. The haptic display used in RH and ARH groups delivered tactile stimuli to the forearm, simulating the bowstring's tension when pulled.

\begin{figure}[!h]
 \centering
\includegraphics[width=1.0\linewidth]{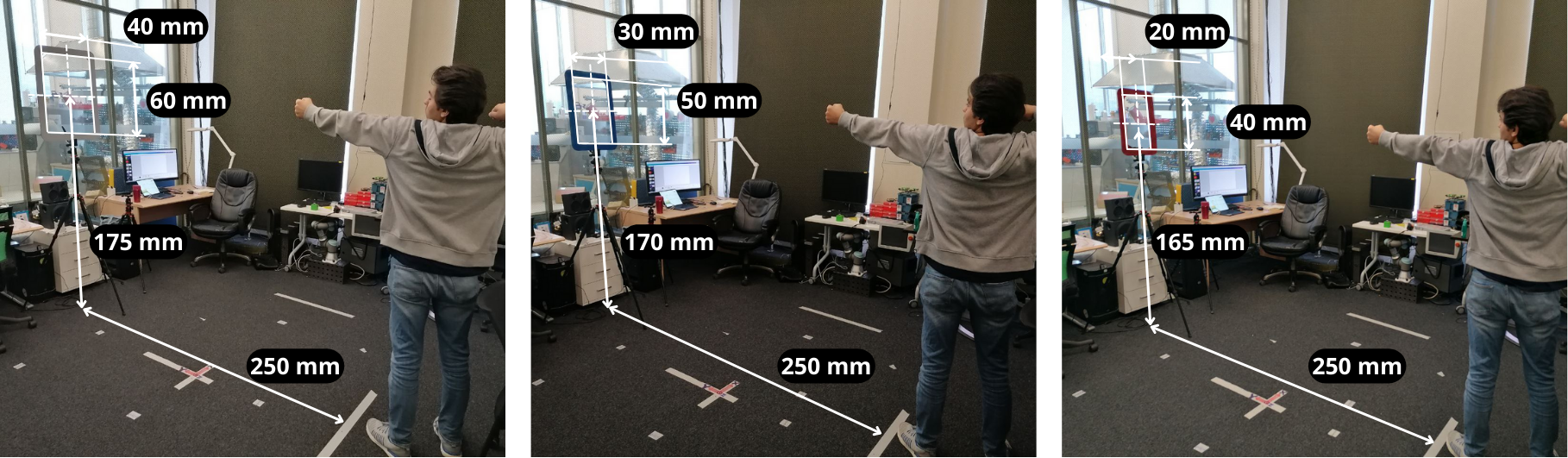}
 \caption{Game elements, i.e. gates, are used as the targets to evaluate the aiming precision.}
 \label{fig:game}
\end{figure}

After the participant had released the stretched virtual bowstring, a shot was fired: the drone stopped following the hand and flew along the parabolic trajectory calculated from the position of the palms.
The gates were located at a distance of 2.5 m from the user; the inner dimensions of the gates are shown in Fig. \ref{fig:game}.

During the game, the number of points scored by users was calculated as evaluation criteria of which group of users was more successful. Additionally, users evaluated their experience of the game on a 5-point Likert scale (Fig. \ref{fig:us2}). For this purpose, four following questions were presented to each user:

\begin{itemize}
 \item Naturalness: How natural were the tension sensations of the virtual bow? (Artificial — Natural)
 \item Target reachability: Was the goal achievable? (Impossible — Achievable)
 \item Physical demand: Was it physically difficult for you to fulfill the game? (Exhausting — Invigorating)
 \item Trajectory predictability: Did the drone fly in the expected direction when released from the virtual bow? (Unpredictable — Expectable)
\end{itemize} 

\subsection{Experimental Results on Conveying the Tangible Experience of Virtual Archery Game}
\label{sec:exp2_res}

\begin{figure}[!h]
 \centering
 \includegraphics[width=0.8\linewidth]{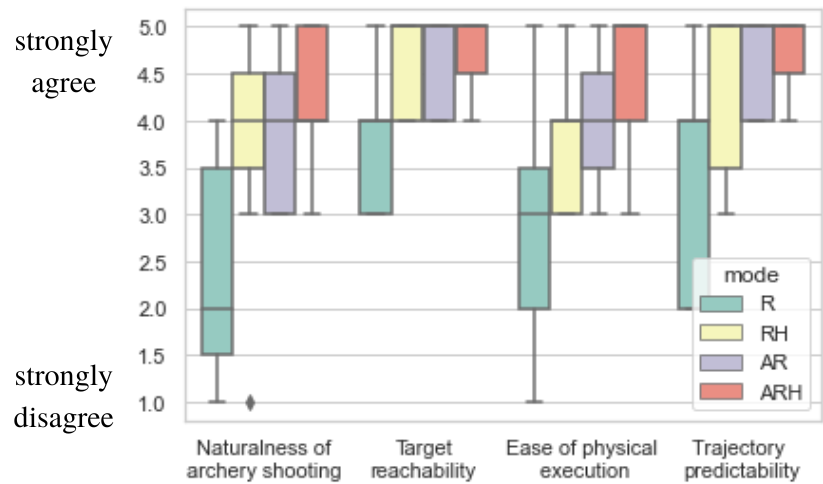}
 \caption{Assessment of the system with a 5-point Likert scale based on user responses.}
 \label{fig:us2}
\end{figure}

The experimental results confirmed that DroneARchery allows users to control the drone accurately and comfortably. Fig. \ref{fig:us2} shows the results of the game assessment by study groups, and these rates are usually higher for the group of users using a tactile display. The results of the game task accomplished by groups are presented in Table \ref{tab:scores}. The average score of the RH group was 12 points, while the participants without a haptic display (R group) achieved an average score of 8 points. The ARH group achieved the highest average score using a haptic display and observing the drone's trajectory in real-time. Therefore, the force estimation accuracy was higher by 22.6\% in the RH group supported with haptic feedback, by 51.5\% in the AR group and 63.3\% in the ARH group compared to the R group with respect to the maximum score (the percentage is obtained by dividing the difference in results for the maximum achievable number of points). The low score in the R and RH categories is the difficulty in predicting the ballistic trajectory for the drone to the target, whereas, in the ARH group, the AR gave the user a visualization of the trajectory, thus aiming at the target more precisely. Additionally, the results revealed that haptic feedback noticeably affected users' performance only when aiming at the smallest gate, with an average result higher by three times in the group supported by LinkTouch. A two-way ANOVA showed statistically significant effect of both gate dimension ($F= 3.46, p=0.034 < 0.05$) and haptic interaction scenario ($F=7.78, p=0.0006 < 0.05$) on the matching score of the players and no significant interaction effect between these parameters: $F = 0.84, p_{int}= 0.499 > 0.05$.

\begin{table}[!ht]
\centering
\caption{Results of the virtual game experiment for 4 groups of users.}
\label{tab:scores}
\scalebox{0.95}{
\begin{tabular}{|m{2.5 cm}| l | l | l | l |}
\hline

Evaluation criteria & R & RH & AR & ARH \\ \hline
Average score grey gate (weighted by 1 point) & 1.3 / 3 & 2.0 / 3 & 2.4 / 3 & 3.0 / 3\\\hline
Average score blue gate (weighted by 3 points) & 0.9 / 3 & 1.6 / 3 & 2.6 / 3 & 2.7 / 3\\\hline
Average score red gate (weighted by 5 points) & 0.6 / 3 & 1.3 / 3 & 2.1 / 3 & 2.6 / 3\\\hline
Total average score & 2.8 / 9 & 4.9 / 9 & 7.1 / 9 & 8.3 / 9 \\\hline
Total average score in points & 7.0 / 27 & 13.1 / 27 & 20.9 / 27 & 24.1 / 27 \\\hline
\end{tabular}
}
\end{table}

Users evaluated the predictability of the final drone trajectory as higher in cases when they observe trajectory, and the haptic display rendered information about the tension of the bow (median of 4.7 in ARH vs 4.5 in AR vs 4.3 in RH vs 3.3 in R). Aside from that, users spend less effort at aiming and setting their hands in the correct position, which affected their fatigue (median of 4.4 in ARH vs 4.0 in AR vs 3.7 in RH vs 2.9 in R). If we consider separately groups using and not using Oculus headset the average value of naturalness in the groups without haptic display is lower than in the groups with tactile feedback (median of 4.3 in ARH vs 3.8 in AR vs 3.7 in RH vs 2.4 in R). For users who did not use AR feedback or a haptic display, the goal was estimated as less achievable (median of 4.7 in ARH vs 4.6 in AR vs 4.6 in RH vs 3.7 in R). Apart from the conclusions about the operation of the system as a whole, a noticeable observation was made, that all participants during the game felt comfortable and calm in the same environment with the drones (4.7 points out of 5.0). 

The two-way ANOVA analysis results showed a statistically significant difference between the Naturalness of archery shooting for four proposed HDI concepts ($F= 3.65, p=0.027 < 0.05$), same as for target reachability ($F= 4.24, p=0.015 < 0.05$), ease of physical execution ($F= 3.08, p=0.046 < 0.05$), trajectory predictability ($F= 3.87, p=0.022 < 0.05$). We also confirmed that the experience of users in interaction with drones does not affect the evaluation according to the chi-square test of independence, for example, trajectory predictability ($\tilde{\chi}^2= 4.43, p=0.61 > 0.05$).

 In addition, we compared the average scores earned by users with and without experience with AR interfaces or drones. There were in total 14 participants in AR and ARH group, where 8 people did not have any experience with AR and drone-based systems. The total average score was 21.0 out of 27 (AR) and 24.6 out of 27 (ARH) for people without experience and 20.7 out of 27 (AR) and 23.8 out of 27 (ARH) for people with experience, supporting our hypothesis that experiences with AR did not significantly affect users' performance. 
 
 \section{DRL Algorithm Evaluation}
 \label{sec:exp3}
 
Drone agents were trained to avoid obstacles over 1000 epochs. Evaluation occurred when interacting with the environment in the simulation where 3 agents avoided collision with each other and with the fourth agent, an arrow. The initial positions of the drones at the beginning of each episode did not change.

\begin{figure*}[!h]
 \centering
 \includegraphics[width=0.95\linewidth]{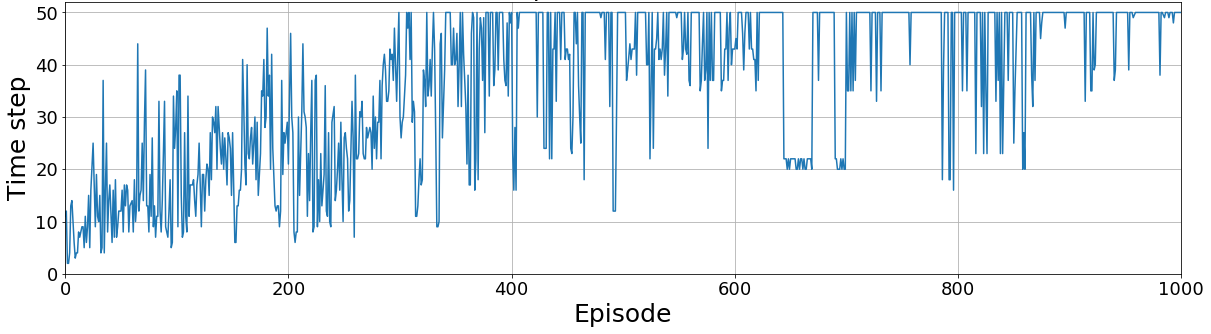} 
 \caption{Episode duration before drones collided during the learning process. The maximum value of 50-time steps means successful performance.}
 \label{fig:time}
\end{figure*}

In the DRL algorithm, the values of hyperparameters leading to the successful completion of the task were selected empirically. The optimal learning rate for the problem being solved is 0.001, the batch size is 10000 states. Hyperparameters $\gamma$ and $\beta$ equals to 0.5 and 0.001 respectively. The execution time and the number of points per episode were taken as metrics. The minimal duration of an episode is 50 timestamps with a 10 Hz frequency, otherwise, drones are colliding with each other (Fig. \ref{fig:time}). Policy and value losses fluctuate greatly over time since we operate in a continuous action space and the speed can take both negative and positive values.

In the simulation, the accuracy of the algorithm was tested during 300 episodes, of which 96\% were successful. In addition, the Reinforcement Learning approach was compared with the Artificial Potential Fields (APF) \cite{khatib1986potential} method to examine the time advantage of calculating a new drone position (average time is 1.35 ms DRL vs 1.24 ms APF for 1000 launches in the simulation). We invited 10 users from the previous user study (5 have experience with drones) to evaluate DRL-based swarm behavior in simulation. The results revealed that users perceived swarm motion as predictable (median 4.4 out of 5) and natural (median 4.0 out of 5).

\section{Conclusions and Future Work}
We developed the DRL-based collision avoidance and drone swarm control for environments with fast-moving obstacles and a new interactive human-drone interaction in AR with a wearable tactile display. Visualized trajectory and haptic feedback allow users to more accurately predict the drone's ballistic flight path, the proposed DRL approach makes interaction safe.

Two user studies were conducted to evaluate users' recognition of haptic patterns, the naturalness of drone deployment in the AR game, and the accuracy of aiming with the help of an augmented environment. The average pattern recognition rate was found to be 68.88\%, suggesting lowering the resolution of sliding distance, as a normal force was recognized at an average of 94.2\%. The accuracy of hitting the target was increased by 63.3\% among users using a haptic display and AR technology. Moreover, the experimental results revealed that users' experience with AR interfaces did not significantly affect their performance in drone launching (score of 24.6 out of 27 for people without experience and 23.8 out of 27 for users with experience). 

The proposed DroneARchery system can be potentially used in the entertainment industry as a novel game scenario. The visualized ballistic trajectory and intuitive haptic interface allow users to set the landing point of the drone at visually occluded areas in a cluttered environment. Additionally, this technology can be implemented in drone teleoperation and swarm teaching of collision avoidance with formation control in presence of dynamic obstacles. Thus, DroneARchery will potentially allow humans and drones to learn from each other through game interaction.
\acknowledgments{
The reported study was funded by RFBR and CNRS, project number 21-58-15006.

}

\bibliographystyle{abbrv-doi}

\bibliography{template}

\begin{thebibliography}{10}

\bibitem{Abdullah_2018}
M.~Abdullah, M.~Kim, W.~Hassan, Y.~Kuroda, and S.~Jeon.
\newblock Hapticdrone: An encountered-type kinesthetic haptic interface with
  controllable force feedback: Example of stiffness and weight rendering.
\newblock In {\em 2018 IEEE Haptics Symposium (HAPTICS)}, pp. 334--339, 2018.
  doi: {{%
10\hspace{.1pt}\discretionary{.}{%
}{.}\hspace{.4pt}1109\discretionary{/}{%
}{/}HAPTICS\hspace{.1pt}\discretionary{.}{%
}{.}\hspace{.4pt}2018\hspace{.1pt}\discretionary{.}{%
}{.}\hspace{.4pt}8357197}}


\bibitem{Byun_2019}
S.-W. Byun and S.-P. Lee.
\newblock Implementation of hand gesture recognition device applicable to smart
  watch based on flexible epidermal tactile sensor array.
\newblock {\em Micromachines}, 10(10), 2019. doi: {{%
10\hspace{.1pt}\discretionary{.}{%
}{.}\hspace{.4pt}3390\discretionary{/}{%
}{/}mi10100692}}


\bibitem{Chen_2020}
M.~Chen, P.~Zhang, Z.~Wu, and X.~Chen.
\newblock A multichannel human-swarm robot interaction system in augmented
  reality.
\newblock {\em Virtual Reality \& Intelligent Hardware}, 2(6):518--533, 2020.
  doi: {{%
10\hspace{.1pt}\discretionary{.}{%
}{.}\hspace{.4pt}1016\discretionary{/}{%
}{/}j\hspace{.1pt}\discretionary{.}{%
}{.}\hspace{.4pt}vrih\hspace{.1pt}\discretionary{.}{%
}{.}\hspace{.4pt}2020\hspace{.1pt}\discretionary{.}{%
}{.}\hspace{.4pt}05\hspace{.1pt}\discretionary{.}{%
}{.}\hspace{.4pt}006}}


\bibitem{Clark_2019}
W.~Clark, M.~Sivan, and R.~O{\textquoteright}Connor.
\newblock Evaluating the use of robotic and virtual reality rehabilitation
  technologies to improve function in stroke survivors: A narrative review.
\newblock {\em Journal of Rehabilitation and Assistive Technologies
  Engineering}, 6, Nov. 2019. doi: {{%
10\hspace{.1pt}\discretionary{.}{%
}{.}\hspace{.4pt}1177\discretionary{/}{%
}{/}2055668319863557}}


\bibitem{Duan_2019}
T.~Duan, P.~Punpongsanon, S.~Jia, D.~Iwai, K.~Sato, and K.~N. Plataniotis.
\newblock Remote environment exploration with drone agent and haptic force
  feedback.
\newblock In {\em 2019 IEEE International Conference on Artificial Intelligence
  and Virtual Reality (AIVR)}, pp. 167--1673, 2019. doi: {{%
10\hspace{.1pt}\discretionary{.}{%
}{.}\hspace{.4pt}1109\discretionary{/}{%
}{/}AIVR46125\hspace{.1pt}\discretionary{.}{%
}{.}\hspace{.4pt}2019\hspace{.1pt}\discretionary{.}{%
}{.}\hspace{.4pt}00034}}


\bibitem{doi:10.3109/08990229109144725}
G.~K. Essick, K.~R. Bredehoeft, D.~F. McLaughlin, and J.~A. Szaniszlo.
\newblock Directional sensitivity along the upper limb in humans.
\newblock {\em Somatosensory \& Motor Research}, 8(1):13--22, 1991. doi: {{%
10\hspace{.1pt}\discretionary{.}{%
}{.}\hspace{.4pt}3109\discretionary{/}{%
}{/}08990229109144725}}


\bibitem{Fedoseev_2020}
A.~Fedoseev, A.~Tleugazy, L.~Labazanova, and D.~Tsetserukou.
\newblock Teslamirror: Multistimulus encounter-type haptic display for shape
  and texture rendering in vr.
\newblock In {\em ACM SIGGRAPH 2020 Emerging Technologies}, SIGGRAPH '20.
  Association for Computing Machinery, New York, NY, USA, 2020. doi: {{%
10\hspace{.1pt}\discretionary{.}{%
}{.}\hspace{.4pt}1145\discretionary{/}{%
}{/}3388534\hspace{.1pt}\discretionary{.}{%
}{.}\hspace{.4pt}3407300}}


\bibitem{gomes2016bitdrones}
A.~Gomes, C.~Rubens, S.~Braley, and R.~Vertegaal.
\newblock Bitdrones: Towards using 3d nanocopter displays as interactive
  self-levitating programmable matter.
\newblock In {\em Proceedings of the 2016 CHI Conference on Human Factors in
  Computing Systems}, pp. 770--780. ACM, 2016.

\bibitem{ibrahimov2019dronepick}
R.~Ibrahimov, E.~Tsykunov, V.~Shirokun, A.~Somov, and D.~Tsetserukou.
\newblock Dronepick: Object picking and delivery teleoperation with the drone
  controlled by a wearable tactile display.
\newblock In {\em 2019 28th IEEE International Conference on Robot and Human
  Interactive Communication (RO-MAN)}, pp. 1--6, 2019. doi: {{%
10\hspace{.1pt}\discretionary{.}{%
}{.}\hspace{.4pt}1109\discretionary{/}{%
}{/}RO\discretionary{%
}{-}{-}MAN46459\hspace{.1pt}\discretionary{.}{%
}{.}\hspace{.4pt}2019\hspace{.1pt}\discretionary{.}{%
}{.}\hspace{.4pt}8956344}}


\bibitem{Juhasz_2008}
T.~Juhász and L.~Vajta.
\newblock Humanoid robot game: A mixture of vr and teleoperation.
\newblock In {\em Advances in Mobile Robotics}, pp. 506--513. doi: {{%
10\hspace{.1pt}\discretionary{.}{%
}{.}\hspace{.4pt}1142\discretionary{/}{%
}{/}9789812835772\_0061}}


\bibitem{kalinov2021warevr}
I.~Kalinov, D.~Trinitatova, and D.~Tsetserukou.
\newblock Warevr: Virtual reality interface for supervision of autonomous
  robotic system aimed at warehouse stocktaking.
\newblock In {\em 2021 IEEE International Conference on Systems, Man, and
  Cybernetics (SMC)}, p. 2139–2145. IEEE Press, 2021. doi: {{%
10\hspace{.1pt}\discretionary{.}{%
}{.}\hspace{.4pt}1109\discretionary{/}{%
}{/}SMC52423\hspace{.1pt}\discretionary{.}{%
}{.}\hspace{.4pt}2021\hspace{.1pt}\discretionary{.}{%
}{.}\hspace{.4pt}9659133}}


\bibitem{swarmplay}
E.~Karmanova, V.~Serpiva, S.~Perminov, A.~Fedoseev, and D.~Tsetserukou.
\newblock Swarmplay: Interactive tic-tac-toe board game with swarm of nano-uavs
  driven by reinforcement learning.
\newblock In {\em 2021 30th IEEE International Conference on Robot \& Human
  Interactive Communication (RO-MAN)}, pp. 1269--1274, 2021. doi: {{%
10\hspace{.1pt}\discretionary{.}{%
}{.}\hspace{.4pt}1109\discretionary{/}{%
}{/}RO\discretionary{%
}{-}{-}MAN50785\hspace{.1pt}\discretionary{.}{%
}{.}\hspace{.4pt}2021\hspace{.1pt}\discretionary{.}{%
}{.}\hspace{.4pt}9515355}}


\bibitem{Kavas_2018}
O.~Kavas and H.~Gurocak.
\newblock Haptic interface with linear magnetorheological (mr) brakes for drone
  control.
\newblock In {\em 2018 15th International Conference on Ubiquitous Robots
  (UR)}, pp. 676--681, 2018. doi: {{%
10\hspace{.1pt}\discretionary{.}{%
}{.}\hspace{.4pt}1109\discretionary{/}{%
}{/}URAI\hspace{.1pt}\discretionary{.}{%
}{.}\hspace{.4pt}2018\hspace{.1pt}\discretionary{.}{%
}{.}\hspace{.4pt}8441784}}


\bibitem{khatib1986potential}
O.~Khatib.
\newblock The potential field approach and operational space formulation in
  robot control.
\newblock In {\em Adaptive and learning systems}, pp. 367--377. Springer, 1986.

\bibitem{RRT}
S.~M. LaValle and J.~James J.~Kuffner.
\newblock Randomized kinodynamic planning.
\newblock {\em The International Journal of Robotics Research}, 20(5):378--400,
  2001. doi: {{%
10\hspace{.1pt}\discretionary{.}{%
}{.}\hspace{.4pt}1177\discretionary{/}{%
}{/}02783640122067453}}


\bibitem{Liu_2020}
C.~Liu and S.~Shen.
\newblock An augmented reality interaction interface for autonomous drone.
\newblock In {\em 2020 IEEE/RSJ International Conference on Intelligent Robots
  and Systems (IROS)}, p. 11419–11424. IEEE Press, 2020. doi: {{%
10\hspace{.1pt}\discretionary{.}{%
}{.}\hspace{.4pt}1109\discretionary{/}{%
}{/}IROS45743\hspace{.1pt}\discretionary{.}{%
}{.}\hspace{.4pt}2020\hspace{.1pt}\discretionary{.}{%
}{.}\hspace{.4pt}9341037}}


\bibitem{48292}
C.~Lugaresi, J.~Tang, H.~Nash, C.~McClanahan, E.~Uboweja, M.~Hays, F.~Zhang,
  C.-L. Chang, M.~Yong, J.~Lee, W.-T. Chang, W.~Hua, M.~Georg, and
  M.~Grundmann.
\newblock Mediapipe: A framework for perceiving and processing reality.
\newblock In {\em Third Workshop on Computer Vision for AR/VR at IEEE Computer
  Vision and Pattern Recognition (CVPR) 2019}, 2019.

\bibitem{Luo_2020}
T.~Luo, N.~Cai, Z.~Li, Z.~Pan, and Q.~Yuan.
\newblock Vr-dlr: A serious game of somatosensory driving applied to limb
  rehabilitation training.
\newblock In N.~J. Nunes, L.~Ma, M.~Wang, N.~Correia, and Z.~Pan, eds., {\em
  Entertainment Computing -- ICEC 2020}, pp. 51--64. Springer International
  Publishing, Cham, 2020.

\bibitem{Macchini_2020}
M.~Macchini, T.~Havy, A.~Weber, F.~Schiano, and D.~Floreano.
\newblock Hand-worn haptic interface for drone teleoperation.
\newblock In {\em 2020 IEEE International Conference on Robotics and Automation
  (ICRA)}, pp. 10212--10218, 2020. doi: {{%
10\hspace{.1pt}\discretionary{.}{%
}{.}\hspace{.4pt}1109\discretionary{/}{%
}{/}ICRA40945\hspace{.1pt}\discretionary{.}{%
}{.}\hspace{.4pt}2020\hspace{.1pt}\discretionary{.}{%
}{.}\hspace{.4pt}9196664}}


\bibitem{Montebaur_2020}
M.~Montebaur, M.~Wilhelm, A.~Hessler, and S.~Albayrak.
\newblock A gesture control system for drones used with special operations
  forces.
\newblock In {\em Companion of the 2020 ACM/IEEE International Conference on
  Human-Robot Interaction}, HRI '20, p.~77. Association for Computing
  Machinery, New York, NY, USA, 2020. doi: {{%
10\hspace{.1pt}\discretionary{.}{%
}{.}\hspace{.4pt}1145\discretionary{/}{%
}{/}3371382\hspace{.1pt}\discretionary{.}{%
}{.}\hspace{.4pt}3378206}}


\bibitem{8357173}
T.~K. Moriyama, A.~Nishi, R.~Sakuragi, T.~Nakamura, and H.~Kajimoto.
\newblock Development of a wearable haptic device that presents haptics
  sensation of the finger pad to the forearm.
\newblock In {\em 2018 IEEE Haptics Symposium (HAPTICS)}, pp. 180--185, 2018.
  doi: {{%
10\hspace{.1pt}\discretionary{.}{%
}{.}\hspace{.4pt}1109\discretionary{/}{%
}{/}HAPTICS\hspace{.1pt}\discretionary{.}{%
}{.}\hspace{.4pt}2018\hspace{.1pt}\discretionary{.}{%
}{.}\hspace{.4pt}8357173}}


\bibitem{DBLP:journals/corr/abs-1910-01806}
E.~Nikonova and J.~Gemrot.
\newblock Deep q-network for angry birds.
\newblock {\em CoRR}, abs/1910.01806, 2019.

\bibitem{9564258}
S.~Ouahouah, M.~Bagaa, J.~Prados-Garzon, and T.~Taleb.
\newblock Deep-reinforcement-learning-based collision avoidance in uav
  environment.
\newblock {\em IEEE Internet of Things Journal}, 9(6):4015--4030, 2022. doi:
  {{%
10\hspace{.1pt}\discretionary{.}{%
}{.}\hspace{.4pt}1109\discretionary{/}{%
}{/}JIOT\hspace{.1pt}\discretionary{.}{%
}{.}\hspace{.4pt}2021\hspace{.1pt}\discretionary{.}{%
}{.}\hspace{.4pt}3118949}}


\bibitem{Ribeiro_2018}
R.~Ribeiro, J.~Ramos, D.~Safadinho, and A.~M. de~Jesus~Pereira.
\newblock Uav for everyone : An intuitive control alternative for drone racing
  competitions.
\newblock In {\em 2018 2nd International Conference on Technology and
  Innovation in Sports, Health and Wellbeing (TISHW)}, pp. 1--8, 2018. doi: {{%
10\hspace{.1pt}\discretionary{.}{%
}{.}\hspace{.4pt}1109\discretionary{/}{%
}{/}TISHW\hspace{.1pt}\discretionary{.}{%
}{.}\hspace{.4pt}2018\hspace{.1pt}\discretionary{.}{%
}{.}\hspace{.4pt}8559538}}


\bibitem{10.1145/3491101.3519908}
Y.~A. Shim, T.~Kim, and G.~Lee.
\newblock Quadstretch: A forearm-wearable multi-dimensional skin stretch
  display for immersive vr haptic feedback.
\newblock In {\em Extended Abstracts of the 2022 CHI Conference on Human
  Factors in Computing Systems}, CHI EA '22. Association for Computing
  Machinery, New York, NY, USA, 2022. doi: {{%
10\hspace{.1pt}\discretionary{.}{%
}{.}\hspace{.4pt}1145\discretionary{/}{%
}{/}3491101\hspace{.1pt}\discretionary{.}{%
}{.}\hspace{.4pt}3519908}}


\bibitem{Tsetserukou_2014}
D.~Tsetserukou, S.~Hosokawa, and K.~Terashima.
\newblock Linktouch: A wearable haptic device with five-bar linkage mechanism
  for presentation of two-dof force feedback at the fingerpad.
\newblock In {\em 2014 IEEE Haptics Symposium (HAPTICS)}, pp. 307--312, 2014.
  doi: {{%
10\hspace{.1pt}\discretionary{.}{%
}{.}\hspace{.4pt}1109\discretionary{/}{%
}{/}HAPTICS\hspace{.1pt}\discretionary{.}{%
}{.}\hspace{.4pt}2014\hspace{.1pt}\discretionary{.}{%
}{.}\hspace{.4pt}6775473}}


\bibitem{SwarmCloak}
E.~Tsykunov, R.~Agishev, R.~Ibrahimov, L.~Labazanova, T.~Moriyama, H.~Kajimoto,
  and D.~Tsetserukou.
\newblock Swarmcloak: Landing of a swarm of nano-quadrotors on human arms.
\newblock In {\em SIGGRAPH Asia 2019 Emerging Technologies}, SA '19, p.
  46–47. Association for Computing Machinery, New York, NY, USA, 2019. doi:
  {{%
10\hspace{.1pt}\discretionary{.}{%
}{.}\hspace{.4pt}1145\discretionary{/}{%
}{/}3355049\hspace{.1pt}\discretionary{.}{%
}{.}\hspace{.4pt}3360542}}


\bibitem{2019}
E.~Tsykunov, R.~Ibrahimov, D.~Vasquez, and D.~Tsetserukou.
\newblock Slingdrone: Mixed reality system for pointing and interaction using a
  single drone.
\newblock {\em 25th ACM Symposium on Virtual Reality Software and Technology},
  Nov 2019. doi: {{%
10\hspace{.1pt}\discretionary{.}{%
}{.}\hspace{.4pt}1145\discretionary{/}{%
}{/}3359996\hspace{.1pt}\discretionary{.}{%
}{.}\hspace{.4pt}3364271}}


\end{thebibliography}
\end{document}